\documentclass[10pt,twocolumn,letterpaper]{article}

\usepackage[pagenumbers]{cvpr} 

\definecolor{cvprblue}{rgb}{0.21,0.49,0.74}
\usepackage[pagebackref,breaklinks,colorlinks,allcolors=cvprblue]{hyperref}
\usepackage{adjustbox} 
\usepackage{pifont}
\usepackage{multirow}
\usepackage{wrapfig}
\usepackage{multicol}
\newcommand{\xmark}{\ding{55}}
\usepackage{colortbl}
\usepackage{subcaption}
\usepackage{longtable} 
\usepackage{amsmath,amssymb,amsfonts}

\def\confName{CVPR}
\def\confYear{2025}

\title{Mamba4D: Efficient 4D Point Cloud Video Understanding with Disentangled Spatial-Temporal State Space Models}

\author{Jiuming~Liu\textsuperscript{\rm 1},
 Jinru~Han\textsuperscript{\rm 1},
 Lihao~Liu \textsuperscript{\rm 2}, Angelica~I. Aviles-Rivero\textsuperscript{\rm 2}\\
 Chaokang~Jiang \textsuperscript{\rm 3},
 Zhe~Liu\textsuperscript{\rm 4},
 Hesheng~Wang \textsuperscript{\rm 1} \thanks{ Corresponding Author. The first two authors contribute equally.}\\
  {\textsuperscript{\rm 1}Department of Automation, Shanghai Jiao Tong University}\qquad
  {\textsuperscript{\rm 2}University of Cambridge}\\
{\textsuperscript{\rm 3}PhiGent Robotics}\qquad
{\textsuperscript{\rm 4} MoE Key Lab of Artificial Intelligence, Shanghai Jiao Tong University}\\
}

\begin{document}
\maketitle
\begin{abstract}

Point cloud videos can faithfully capture real-world spatial geometries and temporal dynamics, which are essential for enabling intelligent agents to understand the dynamically changing world. However, designing an effective 4D backbone remains challenging, mainly due to the irregular and unordered distribution of points and temporal inconsistencies across frames. Also, recent transformer-based 4D backbones commonly suffer from large computational costs due to their quadratic complexity, particularly for long video sequences.
To address these challenges, we propose a novel point cloud video understanding backbone purely based on the State Space Models (SSMs). Specifically, we first disentangle space and time in 4D video sequences and then establish the spatio-temporal correlation with our designed Mamba blocks. The Intra-frame Spatial Mamba module is developed to encode locally similar geometric structures within a certain temporal stride. Subsequently, locally correlated tokens are delivered to the Inter-frame Temporal Mamba module, which integrates long-term point features across the entire video with linear complexity. Our proposed Mamba4d achieves competitive performance on the MSR-Action3D action recognition (+10.4\% accuracy), HOI4D action segmentation (+0.7 F1 Score), and Synthia4D semantic segmentation (+0.19 mIoU) datasets. Especially, for long video sequences, our method has a significant efficiency improvement with 87.5\% GPU memory reduction and $\times$5.36 speed-up. Codes will be released at https://github.com/IRMVLab/Mamba4D.

\end{abstract}   

\begin{figure}[t]
\centering
\includegraphics[width=1.0\linewidth]{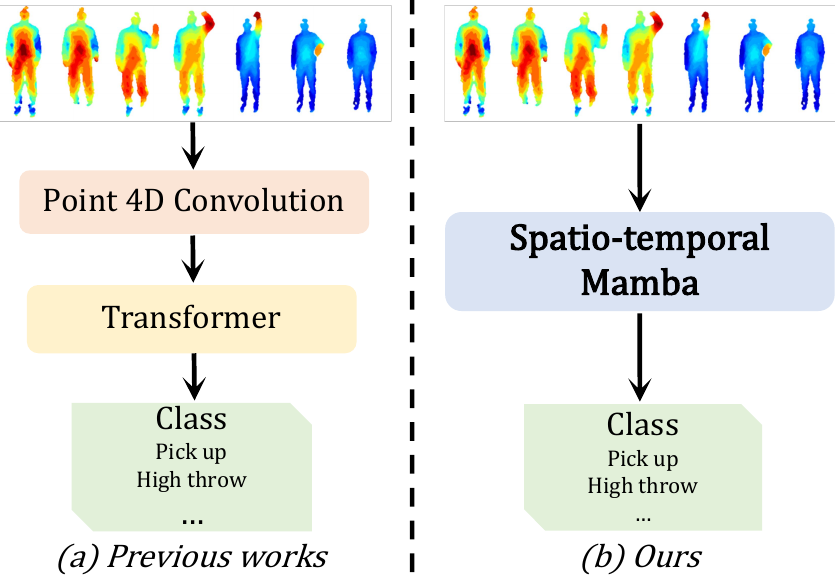}
\vspace{-4mm}
\caption{\textbf{Comparison with previous 4D backbones.} Recent SOTA works \cite{fan2021point,fan2022point,wen2022point} mostly leverage the combination of convolution and transformer to capture the short-term and long-term dynamics, respectively. Our Mamba4D instead utilizes a unified spatio-temporal Mamba module for efficient 4D processing.}
\vspace{-4mm}
\label{fig:head}
\end{figure}
\vspace{-4pt}
\section{Introduction}
\label{sec:intro}
4D point cloud videos, which incorporate 3D space and 1D time, can faithfully represent the geometric structures and temporal motions of our dynamically changing physical world \cite{fan2021point}. They play a crucial role in enabling intelligent agents to perceive dynamics, comprehend environmental changes, and interact effectively with the world \cite{liu2022hoi4d}. Therefore, point cloud video modeling has recently attracted remarkable research interest \cite{wen2022point,shen2023masked,shen2023pointcmp}, with wide-ranging applications, e.g., robotics \cite{deng2023long,zhu2024sni}, augmented and virtual reality (AR/VR) \cite{takagi2000development}, and SLAM systems \cite{liu2023translo,teed2021droid}, etc.


\begin{figure}[t]
\centering
\includegraphics[width=1.0\linewidth]{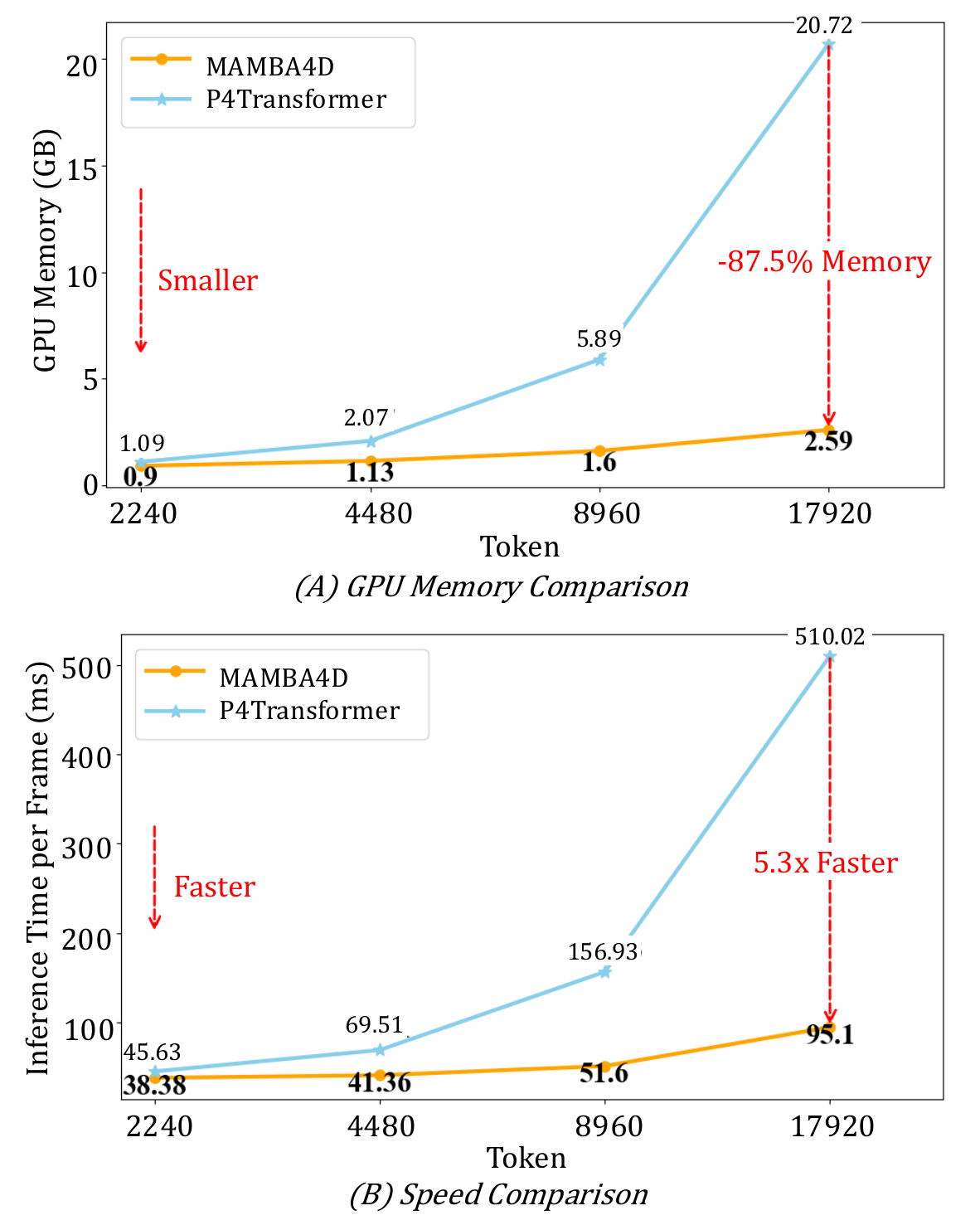}
\vspace{-4mm}
\caption{\textbf{Efficiency comparison with recent SOTA 4D backbones.} We substitute the CNN and transformer backbones in P4Transformer \cite{fan2021point} with our proposed spatio-temporal Mamba models, which leads to \textbf{87.5\% GPU memory reduction} and \textbf{5.36$\times$ faster runtime}. This reveals the great scalability potential of our method for processing long-sequence 4D videos.}
\vspace{-4mm}
\label{fig:compare}
\end{figure}
\vspace{-4pt}

Despite the great success in static 3D point cloud understanding \cite{qi2017pointnet,qi2017pointnet++,choe2022pointmixer,liu2024point}, developing effective backbones for dynamic 4D point cloud sequences remains a significant challenge \cite{wen2022point}. Unlike conventional RGB videos that have structured grids \cite{zheng2020dynamic,ma2024diff}, point cloud videos are inherently irregular and unordered along the spatial dimension \cite{fan2021point}. Therefore, existing backbones and techniques used for grid-based video or static point cloud modeling, e.g., 3D convolutions \cite{karpathy2014large,zheng2023spherical,liu2020psi}, cannot be directly applied. One straightforward method to effectively capture point dynamics is to convert raw 4D sequences into regularly structured voxels \cite{choy20194d,wang20203dv} and then apply 4D convolutions \cite{choy20194d,wang20203dv} on these voxel sequences. However, the voxelization process inevitably leads to quantization errors \cite{fan2021point}. Moreover, modeling all spatio-temporal 4D voxel sequences suffers from inefficiencies, due to the huge computational costs caused by the natural sparsity and large scale of 4D data \cite{choy20194d}.


Another research line \cite{liu2019meteornet,fan2021pstnet,fan2021point,shen2023pointcmp,shen2023masked} avoids the voxelization process and directly processes raw 4D sequences, which implicitly encodes 4D spatio-temporal information using convolution networks \cite{fan2021pstnet,wang2022pointmotionnet} or transformers \cite{fan2021point,shen2023pointcmp,shen2023masked,liu2023scotch}. However, the locally restrained receptive field of CNN limits the long-sequence modeling ability \cite{fan2021point,wen2022point}. Recent SOTA methods commonly combine the 4D CNN and transformer as in Fig. \ref{fig:head}, which suffer from high computational burdens due to the quadratic complexity \cite{wen2022point} of the transformer.

To address the efficiency and receptive field challenges, we resort to the Mamba model \cite{gu2022efficiently,gu2023mamba} in this work, because of the linear computational complexity and long-range dependency modeling abilities of the State Space Models (SSMs). Drawing inspirations from previous 4D backbones, we have two principal observations: (i) Point cloud sequences are irregular and unordered along the spatial dimension but regular along the temporal dimension \cite{fan2021pstnet}. Therefore, disentangling orthogonal space and time dimensions is beneficial for effective 4D understanding \cite{fan2021point,fan2022point}, as it reduces the effects of spatial irregularity on temporal modeling. (ii) Multiple granularity information \cite{shen2023masked,shen2023pointcmp} is crucial in modeling spatio-temporal correlation. Local spatio-temporal modeling allows for the association of more fine-grained patterns with similar or related geometric details over time. On the contrary, global temporal correlation enhances higher-level semantic understanding of 4D sequences \cite{liu2023regformer}, such as identifying consistent actions like `a person raising a hand'.



Motivated by the aforementioned observations, we design an intra-frame Mamba module and an inter-frame Mamba module to model short-term and long-term spatio-temporal correlations, respectively. Specifically, we first partition raw 4D sequences into a series of short-term video clips around the stride-based sampling anchor frames. Then, spatio-temporal neighbor points within clips are integrated by the combination of an Intra-frame Spatial Mamba module and a cross-frame temporal pooling layer, which is tailored to capture short-term local structures effectively. To address long-range dependencies along the temporal dimension, we employ the Inter-frame Temporal Mamba block, which receives all locally encoded tokens from all anchor frames. Finally, the spatio-temporally correlated features are fed into different heads tailored for various downstream tasks. Compared with previous transformer counterparts, our Mamba-based 4D backbone excels significantly in efficiency as in Fig. \ref{fig:compare}, which reveals the great scalability potential for processing long-sequence 4D videos.

Overall, our main contributions are as follows:
\begin{itemize}
\item We propose a novel 4D backbone for generic point cloud video understanding purely based on the state space model, termed Mamba4D. To the best of our knowledge, this is the first Mamba-based generic 4D backbone, taking advantages of both long-range dependencies modeling capability and linear complexity.
\item Spatio-temporally decoupled Mamba modules are designed to capture the multiple granularity spatio-temporal correlation. An Intra-frame Spatial Mamba with a temporal pooling layer is developed to encode local structures and short-term dynamics. To establish the long-term spatio-temporal association, another Inter-frame Temporal Mamba is also proposed.
\item Various spatio-temporal scanning strategies are systematically compared in our Inter-frame Temporal Mamba to better correlate long-sequence dynamics. 
\item Experiment results demonstrate the superiority of our Mamba4D on various 4D understanding tasks, including MSR-Action3D action recognition, HOI4D action segmentation, and Synthia4D semantic segmentation. Compared with the transformer-based counterparts, our Mamba4D has significantly enhanced efficiency, empowering the efficient long-sequence video understanding.
\end{itemize}

\section{Related work}

\textbf{CNN-based 4D understanding backbones.} Early works convert 4D points to structured representations, e.g. voxels \cite{choy20194d,wang20203dv}, BEV \cite{luo2018fast}, etc, and then leverage grid based convolutions. However, the transformation process inevitably results in the loss of geometric details. To better retain the raw geometric information, recent CNN-based methods directly process 4D raw sequences. PointRNN \cite{fan2019pointrnn} introduces recurrent neural networks for moving point cloud processing. Meteornet \cite{liu2019meteornet} appends an additional temporal dimension (1D) to 3D points and strengthens PointNet++ \cite{qi2017pointnet++} for tracking points with a grouping strategy or scene flow estimator \cite{liu2019flownet3d,liu2024difflow3d,wang2022sfgan,jiang20243dsflabelling,liu2023traffic}. Nonetheless, the explicit point tracking performance is significantly influenced by the motion inconsistency, since dynamic points may flow in and flow out across consecutive frames \cite{fan2021pstnet}. To overcome this, most subsequent works \cite{fan2021pstnet,fan2021point,wang2022pointmotionnet} implicitly encode spatio-temporal information. As a representative work, PST-Net \cite{fan2021pstnet} designs the temporal and spatial convolution networks. However, the receptive field of CNNs is restrained locally, which significantly undermines the long-sequence temporal modeling \cite{fan2021point}. Therefore, recent SOTA methods resort to transformers for enlarged receptive fields.

\vspace{3pt}
\noindent\textbf{Transformer-based 4D understanding backbones.} P4Transformer \cite{fan2021point} first reveals the great potential of the transformer backbone for 4D understanding tasks, which combines a Point 4D Convolution layer for local structures and a temporal modeling transformer over all timestamps. PPTr \cite{wen2022point} tackles the efficiency constraints of P4Transformer by leveraging primitive plane as a compact mid-level representation. However, these supervised networks require expensive and laborious 4D labels. Therefore, recent transformer-based methods resort to self-supervised learning with 4D knowledge distillation \cite{zhang2023complete}, contrastive mask prediction \cite{shen2023pointcmp,shen2023masked,liu2023contrastive}, or introducing local rigid motion \cite{liu2023leaf}. However, due to the quadratic complexity, existing transformer-based methods commonly struggle with high computational overhead. 


\vspace{3pt}
\noindent\textbf{State space models.} Recently, there has been a significantly increasing focus on the state space models (SSMs) \cite{gu2022efficiently,gu2023mamba}, which have competitive long-range dependency modeling ability compared with transformers. The crucial advantage of SSMs lies in the linear scaling capability, providing a global receptive field with linear computational complexity. HiPPO \cite{gu2020hippo} draws inspiration from the control system and performs online compression of continuous and discrete signals by the polynomial projection. Based on the HiPPO initialization, LSSL \cite{gu2021combining} achieves long-sequence modeling performance but struggles with prohibitive computation and memory requirements. To overcome this problem, S4 \cite{gu2022efficiently} introduces a Structured State-Space Sequence model and makes SSMs feasible to be used in practical tasks like speech and language. However, S4 cannot perform well in terms of selective copying and induction heads. Recently, Mamba \cite{gu2023mamba} proposes a data-dependent SSM block with a parallel scan algorithm and hardware-aware kernel fusion mechanism. To this end, state space models have evolved into a relatively mature stage.

\textbf{Mamba for vision applications.} Mamba has opened a new era for various promising vision backbones, including 2D image \cite{zhu2024vision,liu2024vmamba,ma2024u,xing2024segmamba}, 3D point cloud \cite{liang2024pointmamba,zhang2024point,liu2024point}, RGB video \cite{yang2024vivim,li2024videomamba,chen2024video}, and visual content generation \cite{zhang2024motion,oshima2024ssm}. However, the great potential of Mamba in the 4D point cloud video processing is still unexplored. We notice that recent work Simba \cite{chaudhuri2024simba} proposes a Mamba-augmented U-ShiftGCN network for skeletal action recognition task. Distinct from them, our Mamba4D aims to build a pure SSM-based generic 4D vision backbone for various downstream tasks, e.g. human action recognition, action segmentation, 4D semantic segmentation, etc.



\begin{figure*}[t]
\centering
\includegraphics[width=1.0\linewidth]{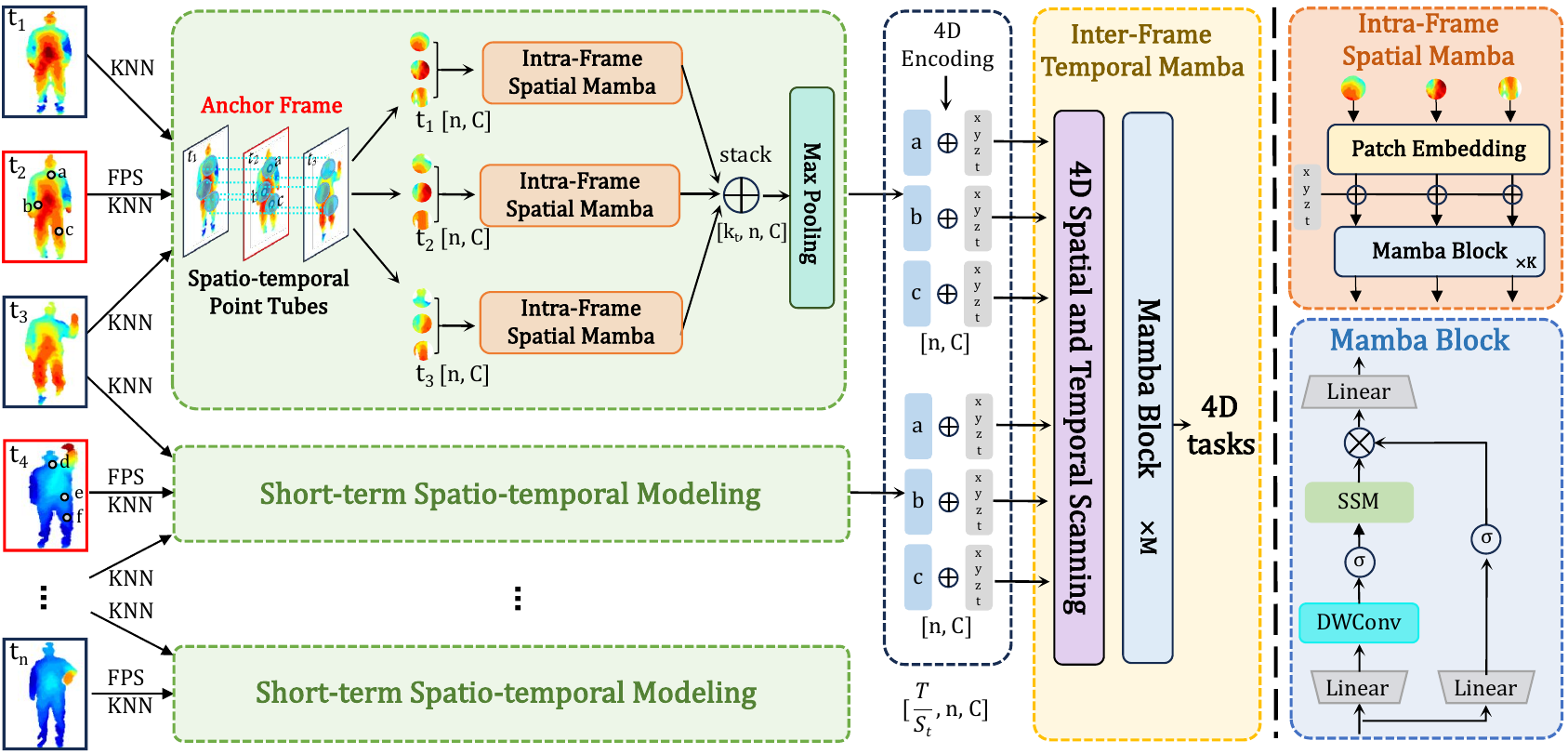}
\vspace{-4mm}
\caption{\textbf{The overview of our Mamba4D.} To capture hierarchical 4D video dynamics, we design an Intra-frame Spatial Mamba on short-term video clips for local dynamic structures and an Inter-frame Temporal Mamba on the entire video sequence for global video understanding. Various spatio-temporal scanning strategies are proposed to better establish 4D correlation.}
\vspace{-4mm}
\label{fig:pipeline}
\end{figure*}


\section{Methodology}
In this section, we delve into the detailed designs of our proposed Mamba4D. The overall pipeline of our proposed method is shown in Fig. \ref{fig:pipeline}, which is mainly composed of an Intra-frame Spatial Mamba and an Inter-frame Temporal Mamba. Specifically, for capturing hierarchical temporal dynamics, some anchor frames are first selected and the whole point cloud sequence is divided into a series of short clips around each anchor frame as in Section \ref{sec:anchor}. For each frame within the same clip, K Nearest Neighbors (KNN) are chosen and then spatio-temporal local structures are captured by the combination of an Intra-frame Spatial Mamba and a short-term temporal pooling layer in section \ref{sec:intra}. To establish the long-range temporal dependencies, an Inter-frame Temporal Mamba is designed as in Section \ref{sec:inter}.

\subsection{Anchor Frame-based 4D Video Partition}
\label{sec:anchor}
Motivated by previous works \cite{fan2021pstnet,fan2021point,zhang2023complete}, multi-granularity information (both short-term and long-term dynamics) is extremely crucial for the accurate spatio-temporal correlation. Therefore, we first select some anchor frames to partition the input video sequence into a series of point tubes \cite{shen2023masked}, on which short-term spatio-temporal dynamics are then captured by an Intra-frame Mamba module and a temporal max-pooling layer. 

\vspace{3pt}
\noindent\textbf{Temporal anchor frame selection.} A stride-based sampling strategy \cite{fan2021pstnet} is adopted along the temporal dimension to generate the anchor frames. In detail, given a point cloud video \{[$P_{t},F_{t}$], $P_{t} \in \mathbb{R}^{N\times 3}$, $F_{t} \in \mathbb{R}^{N\times C}$\}$^{T}_{t=1}$, anchor frames are chosen based on the temporal stride $s_{t}$. $P_{t} \in \mathbb{R}^{N\times 3}$ and $F_{t} \in \mathbb{R}^{N\times C}$ respectively indicate the point coordinates and features for the $t$-th frame. $T$ is the video length. $N$ is the number of points for each frame. Then, the entire video sequence is divided into $\frac{T}{s_{t}}$ equal-length clips with a temporal kernel size $k_{t}$ around each anchor frame.

\textbf{Spatial anchor point selection.} After getting anchor frames, we employ Farthest Point Sampling (FPS) \cite{qi2017pointnet++} to sample spatial anchor points. With a spatial downsampling rate $s_{s}$, $n=\frac{N}{s_{s}}$ anchor points are obtained for each anchor frame. The 3D coordinates of anchor points are then propagated to neighboring frames within the same video clip \cite{fan2021pstnet}. Spatio-temporal K nearest neighbor points (KNN), which are within a certain spatial kernel $k_{s}$ around the anchor points, are selected in the anchor frame and its neighboring frames. Overall, for each anchor point, its neighboring points across the short-term temporal clip will form tube-like spatio-temporal intervals, termed point tubes \cite{shen2023masked}.

\begin{figure*}[t]
\centering
\includegraphics[width=1.0\linewidth]{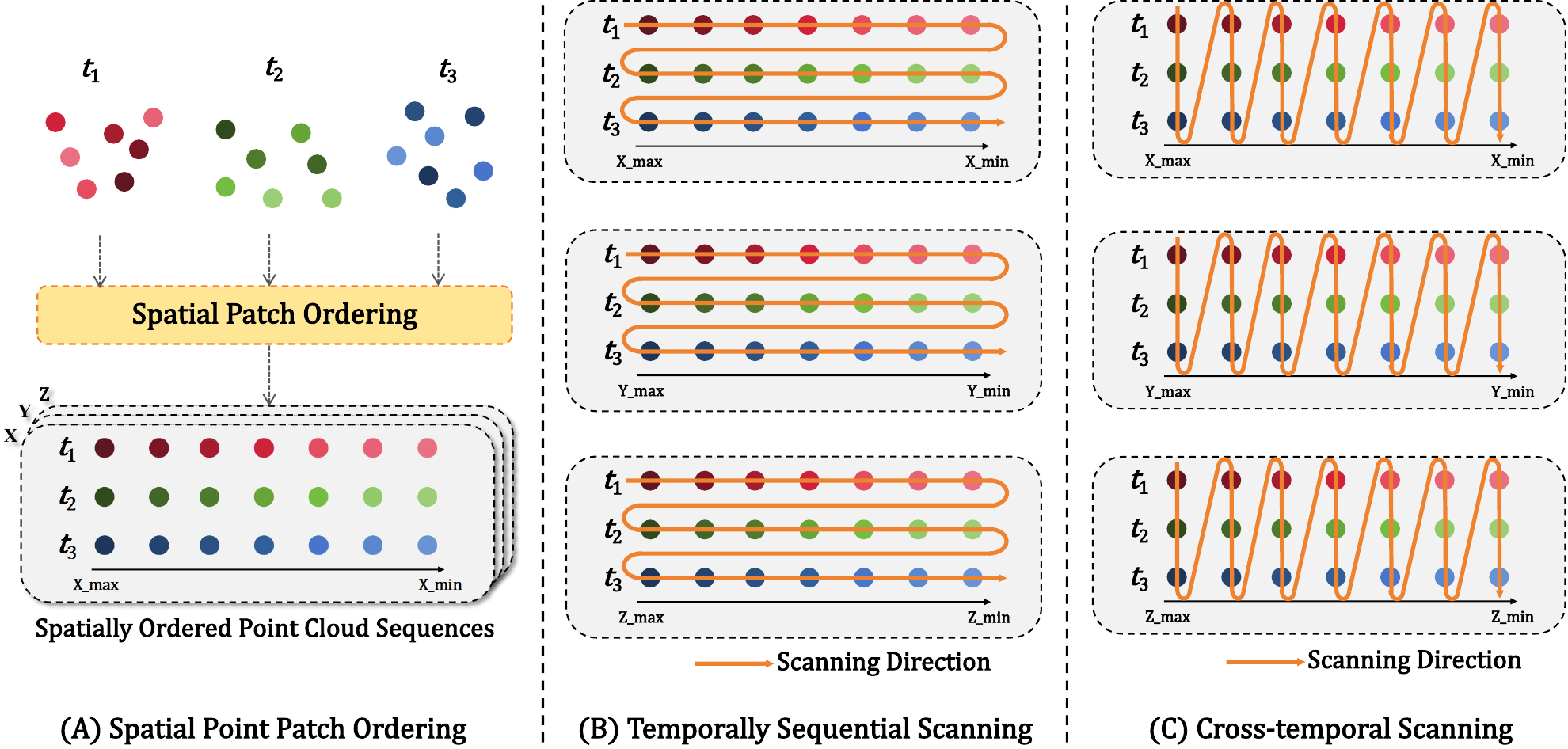}
\vspace{-4mm}
\caption{\textbf{Different spatio-temporal ordering strategies in Inter-frame Temporal Mamba.} We first spatially order all the input point frames according to the X, Y, and Z coordinates. Then, point tokens are scanned by temporal sequences or cross-temporal sequences.}
\vspace{-4mm}
\label{fig:order}
\end{figure*}

\subsection{Intra-frame Spatial Mamba}
\label{sec:intra}
Once short-term spatio-temporal point tubes are formed, the core issue is how to effectively gather local spatio-temporal features within these point tubes. Specifically, spatial information is first independently encoded within each frame by an intra-frame Mamba. Here, we follow \cite{liang2024pointmamba} to design our intra-frame Mamba consisting of $K$ Mamba blocks. For each Mamba block, layer normalization (LN), SSM, depth-wise convolution (DW) \cite{chollet2017xception}, and residual connections are leveraged as:
\begin{equation}
    \label{eq:9}
     F^{'}_{l} = DW(MLP(LN(F_{l-1}))),
\end{equation}
\begin{equation}
    \label{eq:10}
     F_{l} = MLP(LN(SSM(\sigma(F^{'}_{l}))) \times \sigma(LN(F_{l-1}))) + F_{l-1},
\end{equation}
where $F_{l}$ is the output of the $l$-th block. $F_{0}$ is the input features of all point patches from KNN searching. $\sigma(\cdot)$ indicates the SiLU activation function \cite{hendrycks2016gaussian}. 


To better establish the spatio-temporal correlation, the input tokens of the Mamba block should not only have the 3D coordinate encoding but also the awareness of which frame the point comes from. Therefore, we design a 4D positional encoding $PE = MLP(x,y,z,t)$ derived from spatial dimensions $x,y,z$ and temporal dimension $t$. We feed the 4D encoding together with point tokens into our Mamba block. Finally, a temporal Max-pooling layer is utilized to integrate the short-term temporal information within each point cube. Overall, the short-term spatio-temporal modeling module in Fig. \ref{fig:pipeline} can be represented as:
\begin{equation}
\resizebox{\columnwidth}{!}{%
$\begin{aligned}
    \label{eq:11}
     F^{'(x,y,z)}_{t} &= \sum_{(\delta_{x}, \delta_{y}, \delta_{z}, \delta_{t}) \in C}\gamma(\delta_{x}, \delta_{y}, \delta_{z}, \delta_{t}) \circ F^{(x+\delta_{x},y+\delta_{y},z+\delta_{z})}_{t+\delta_{t}} \\
     &= \sum_{\delta_{t}= {-\lfloor{\frac{k_{t}}{2}}\rfloor}}^{\lfloor{\frac{k_{t}}{2}}\rfloor}  \sum_{||\delta_{x}, \delta_{y}, \delta_{z}||\leq k_{s}}  \gamma(\delta_{x}, \delta_{y}, \delta_{z}, \delta_{t}) \circ F^{(x+\delta_{x},y+\delta_{y},z+\delta_{z})}_{t+\delta_{t}}\\
     &= \sum_{\delta_{t}= {-\lfloor{\frac{k_{t}}{2}}\rfloor}}^{\lfloor{\frac{k_{t}}{2}}\rfloor} \mathbf{T}^{(\delta_{t})} \circ \sum_{||\delta_{x}, \delta_{y}, \delta_{z}||\leq k_{s}}  \mathbf{S}^{(\delta_{x}, \delta_{y}, \delta_{z})} \circ  F^{(x+\delta_{x},y+\delta_{y},z+\delta_{z})}_{t+\delta_{t}},
\end{aligned}$
}
\end{equation}
where $(x,y,z)\in P_{t}$ are the coordinates of anchor points. $(\delta_{x}, \delta_{y}, \delta_{z}, \delta_{t})$ indicate the spatio-temporal displacements along the spatial and temporal dimensions in our designed point cube $C$. $\mathbf{S}^{(\delta_{x}, \delta_{y}, \delta_{z})}\circ()$ and $\mathbf{T}^{(\delta_{t})}\circ()$ respectively represent the Intra-frame Spatial Mamba and the temporal pooling layers.

\subsection{Inter-frame Temporal Mamba}
\label{sec:inter}
To model the long-range dependencies for global dynamics, we develop an inter-frame Mamba block on the above locally correlated point tokens from all video clips to gather long-range temporal correlation. Inspired by the recent advancements in 2D and 3D Mambas \cite{zhu2024vision,liu2024vmamba,liang2024pointmamba,zhang2024point} where the scanning mechanism is important for causal orderings, we design two scanning strategies here to adequately model the contextual long-term temporal information: Temporally Sequential Scanning and Cross-temporal Scanning as in Fig. \ref{fig:order}. Specifically, point clouds are first spatially ordered in terms of XYZ coordinates respectively. Afterwards, two scanning strategies are introduced. For the Temporally Sequential Scanning, spatio-temporal points are scanned within each frame first and then scanned along the temporal dimension. For the Cross-temporal Scanning, points with the largest X coordinate are scanned for each frame until the smallest ones. For each scanning strategy, one-time orderings (X, Y, Z) and three-time orderings (XYZ, XZY, YXZ, YZX, ZXY, ZYX) can be adopted which are systematically compared in Section \ref{sec:ablation}. 



\section{Experiment}
\label{exp}
\subsection{3D Action Recognition}
\textbf{Dataset and settings.} We first evaluate our Mamba4D on the MSR-Action3D dataset \cite{li2010action} to demonstrate the effectiveness for 3D action recognition, a video-level classification task. We follow P4Transformer \cite{fan2021point} to partition the training and testing sets. The temporal kernel $k_{t}$ is set as 3, and the temporal stride $s_{t}$ is set as 2. In KNN, spatial stride $s_{s}$ is set as 32, and $k_{s}$ is set as 0.7 for the spatial searching range. The Mamba block number of the intra-frame Mamba \cite{liang2024pointmamba} is 12, and the Mamba block number of the inter-frame Mamba is 4. We train our model on a single NVIDIA RTX 4090 GPU for 50 epochs with the SGD optimizer, where the initial learning rate is set as 0.01, and decays with a rate of 0.1 at the 20-th epoch and the 30-th epoch respectively.

\vspace{3pt}
\noindent\textbf{Comparison results.} As shown in Table \ref{tab:MSR}, we evaluate our model and compare with both CNN-based \cite{qi2017pointnet++,liu2019meteornet,fan2021pstnet} and Transformer-based \cite{fan2021point,shen2023masked} counterparts. From the table, our method consistently outperforms CNNs and transformers by a large margin. Compared with our baseline P4Transformer \cite{fan2021point}, our Mamba4D has higher recognition accuracy for all 24, 32, and 36 frames' inputs. Particularly, with the frame length increasing, our baseline P4Transformer \cite{fan2021point} has an obvious performance drop, which indicates poor modeling for long-sequence 4D videos. On the contrary, the recognition accuracy of our method is steadily improved with much longer sequences. Furthermore, the efficiency of our Mamba4D outperforms our baseline P4Transformer dramatically in terms of both GPU memory (87.5\% reduction) and inference time ($\times$5.36 faster) as in Fig. \ref{fig:compare}. This demonstrates the great potential for the scalability of longer input tokens.

\begin{table}[!t]
\centering
\begin{minipage}[t]{0.50\textwidth}
            \caption{\textbf{Evaluation for action recognition on MSR-Action3D dataset \cite{li2010action}.} We establish our Mamba4D based on recent representative transformer backbones \cite{fan2021point, fan2022point}.}
            \label{tab:MSR}
            \resizebox{\linewidth}{!}{%
            \begin{tabular}{c|c|c|c}
            \hline \toprule
            \cellcolor[HTML]{EFEFEF}Backbone & \cellcolor[HTML]{EFEFEF}Method & \cellcolor[HTML]{EFEFEF}Frame & \cellcolor[HTML]{EFEFEF}Accuracy (\%) \\
            \midrule
            &PointNet++ \cite{qi2017pointnet++} & 1 & 61.61 \\
            &MeteorNet \cite{liu2019meteornet}

            & 24 & 88.50 \\
            \multirow{-3}{*}{\begin{tabular}[c]{@{}c@{}} CNN\end{tabular}}& PSTNet \cite{fan2021pstnet}
            & 24 & 91.20 \\
            \midrule
            &PPTr \cite{wen2022point}

            & 24 & 92.33 \\
             \cmidrule{2-4}
              &MaST-Pre \cite{shen2023masked}
              & 24 & 91.29 \\
              \cmidrule{2-4}

            & & 24 & 90.94 \\
            &  & 32 & 87.93 \\
            \multirow{-5}{*}{\begin{tabular}[c]{@{}c@{}} Transformer\end{tabular}}& \multirow{-3}{*}{\begin{tabular}[c]{@{}c@{}} P4Transformer \cite{fan2021point} \end{tabular}} & 36 & 82.81 \\
             \cmidrule{2-4}
             & PST-Transformer \cite{fan2022point} & 24 & 93.03 \\

            \midrule

            &  & 24 & \textbf{92.68} \\
           &  & 32 & \textbf{93.10} \\
          \multirow{-2}{*}{\begin{tabular}[c]{@{}c@{}} Mamba\end{tabular}}& \multirow{-3}{*}{\begin{tabular}[c]{@{}c@{}} P4Transformer + Ours\end{tabular}} & 36 & \textbf{93.23} \\
             \cmidrule{2-4}
             & PST-Transformer + Ours & 24 & \textbf{93.38} \\

            \bottomrule
    \end{tabular}}
\end{minipage}
\end{table}

\vspace{3pt}
\noindent\textbf{Generalization on other baselines.} Although recent works \cite{wen2022point,fan2022point,zhang2023complete,shen2023pointcmp} surpass P4Transformer in accuracy with their specific carefully-designed strategies, both Point 4D Convolution and transformer from P4Transformer are commonly adopted. Here, we also introduce our proposed method to a more recent SOTA method PST-Transformer \cite{fan2022point} by replacing the Point 4D convolution with our designed Intra-frame Spatial Mamba block as in Fig. \ref{fig:compare}, which proves the universality of our Mamba-based backbone.

\subsection{4D Action Segmentation}
\textbf{Dataset and settings.}
We further conduct the experimental evaluation on the action segmentation task on the HOI4D dataset \cite{liu2022hoi4d}. The objective is to assign action labels to individual frames within point cloud videos. The dataset follows the official HOI4D split, comprising 2971 scenes for training and 892 scenes for testing. Each sequence consists of 150 frames, with each frame containing 2048 points. We utilized X4D-SceneFormer \cite{jing2024x4d}, the current state-of-the-art model, as our backbone architecture. Performance evaluation was conducted using multiple metrics: frame-level accuracy (Acc), segmental edit distance, and segmental F1 scores. The F1 scores were calculated at three different overlapping thresholds: 10\%, 25\%, and 50\%, which are based on Intersection over Union (IoU) ratios \cite{zhang2023complete}.

\vspace{3pt}
\noindent\textbf{Comparison results.} We establish Mamba blocks on a recent state-of-the-art method X4D-SceneFormer \cite{jing2024x4d}. As in Table \ref{tab:action_segmentation}, our Mamba4D surpasses all existing 4D backbones and leads the board on the open-resource HOI4D website.

\begin{figure}[t]
\centering
\includegraphics[width=1.0\linewidth]{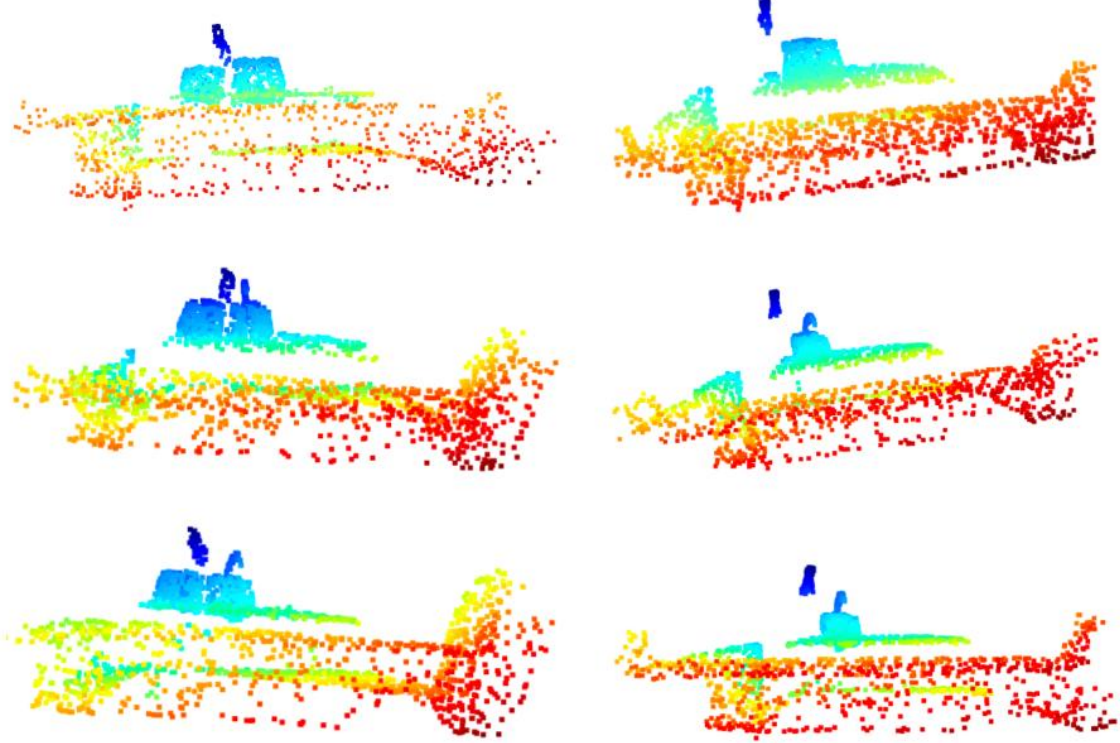}
\vspace{-6mm}
\caption{\textbf{Visualization of the action segmentation on HOI4D.} Here, we display consecutive frame sequences when a person picks and places a mug on the HOI4D dataset.}
\vspace{-2mm}
\label{fig:visual2}
\end{figure}

\vspace{3pt}
\noindent\textbf{Visualization samples.} We visualize one sample of consecutive action sequences in Fig. \ref{fig:visual2} with segmented labels, where a person picks and places a mug.

\begin{table}[!t]
\centering
\caption{\textbf{Action segmentation on the HOI4D dataset \cite{liu2022hoi4d}.}}
\label{tab:action_segmentation}
\resizebox{1.00\columnwidth}{!}
            {
\begin{tabular}{l|c|c|c|c|c}
\toprule
\cellcolor[HTML]{EFEFEF} Method  & \cellcolor[HTML]{EFEFEF}Acc & \cellcolor[HTML]{EFEFEF}Edit & \cellcolor[HTML]{EFEFEF}F1@10 & \cellcolor[HTML]{EFEFEF}F1@25 & \cellcolor[HTML]{EFEFEF}F1@50 \\
\midrule
P4Transformer \cite{fan2021point}  & 71.2 & 73.1 & 73.8 & 69.2 & 58.2 \\
P4Transformer + C2P \cite{zhang2023complete}  & 73.5 & 76.8 & 77.2 & 72.9 & 62.4 \\
PPTr \cite{wen2022point}  & 77.4 & 80.1 & 81.7 & 78.5 & 69.5 \\
PPTr+C2P \cite{zhang2023complete}  & 81.1 & 84.0 & 85.4 & 82.5 & 74.1 \\
X4D-SceneFormer \cite{jing2024x4d}  & 84.1 & 91.1 & 92.5 & 90.8 & 84.8 \\
Ours  & \textbf{85.5} & \textbf{91.3} & \textbf{92.6} & \textbf{91.2} & \textbf{85.5} \\
\bottomrule
\end{tabular}}
\end{table}

\subsection{4D Semantic Segmentation}
\textbf{Dataset and settings.} We further test our method on Synthia 4D dataset \cite{choy20194d} for 4D semantic segmentation. For the semantic segmentation task, additional temporal information is beneficial to improve segmentation accuracy and robustness for noise and occlusion. 

\vspace{3pt}
\noindent\textbf{Comparison results.} We compare our Mamba4D with CNN-based and transformer-based counterparts in Table \ref{tab:synthia}. Without any specific designs, our method consistently outperforms previous CNNs and transformers on most sub-sequences. Compared with our baseline \cite{fan2021point}, our Mamba4D has a 0.19 mIoU improvement. 

\begin{figure*}[t]
\centering
\includegraphics[width=0.8\linewidth]{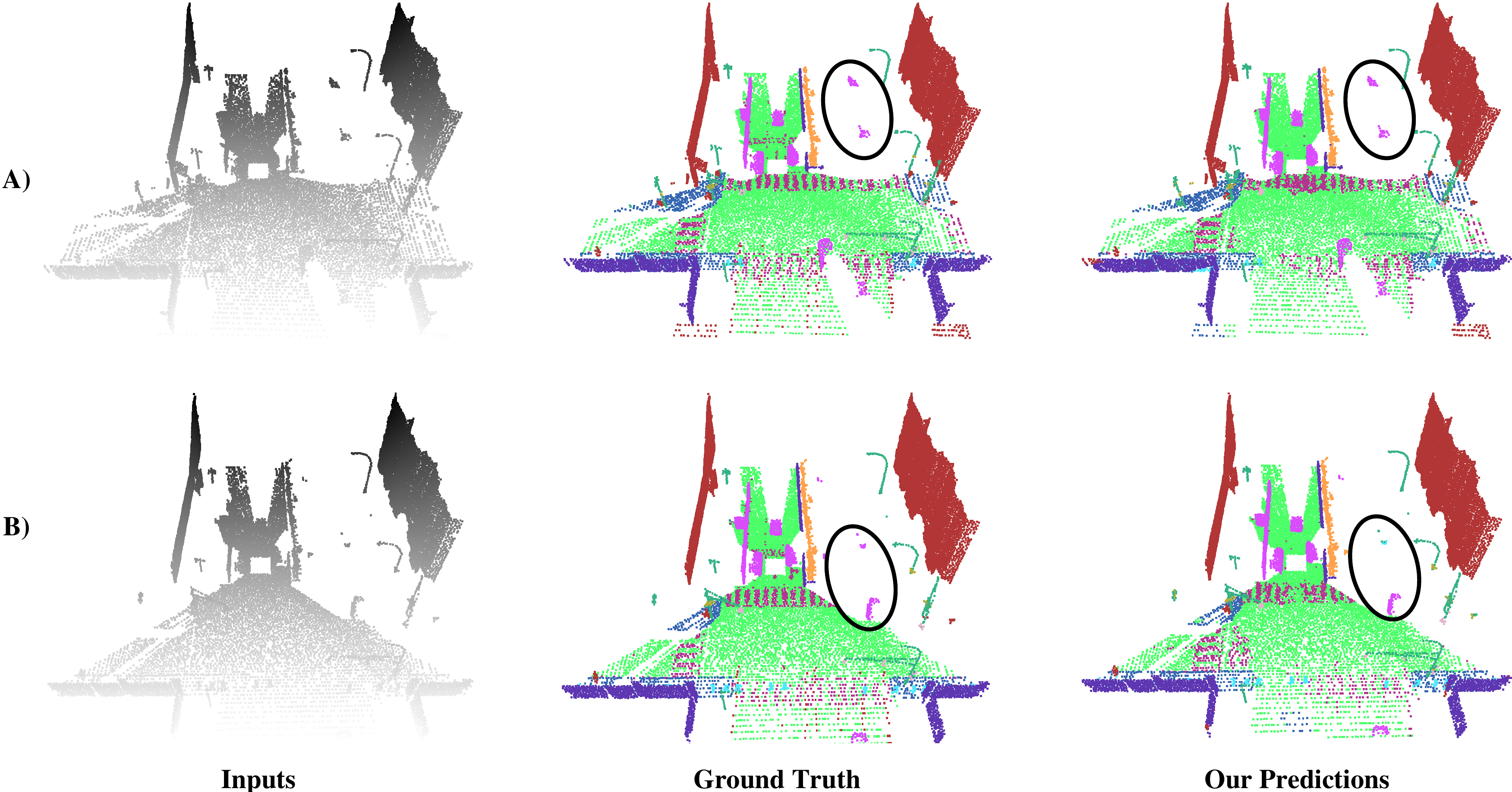}
\vspace{-2mm}
\caption{\textbf{Visualization of the 4D semantic segmentation.} Our method can effectively segment semantic labels in dynamic scenes. We highlight the dynamic cars in the 4D point cloud sequence with a black circle.}
\label{fig:visual}
\end{figure*}

\vspace{3pt}
\noindent\textbf{Visualization samples.} We also display two visualization samples of our segmentation results in Fig. \ref{fig:visual}. Our method can accurately segment the semantic classes in dynamic scenes, which can demonstrate the effectiveness of our method.

\setlength{\tabcolsep}{3mm}
\begin{table*}[t]
\centering
\caption{\textbf{4D semantic segmentation results (mIoU \%) on the Synthia 4D dataset \cite{choy20194d}.} We replace the CNN and transformer architectures in P4Transformer \cite{fan2021point} with our proposed Mamba blocks across different frames.}
\label{tab:synthia}
\resizebox{\linewidth}{!}{
\begin{tabular}{c|c|c|c|ccccccc}
\toprule
\cellcolor[HTML]{EFEFEF}Method & \cellcolor[HTML]{EFEFEF}Input & \cellcolor[HTML]{EFEFEF}Frame & \cellcolor[HTML]{EFEFEF}Track & \cellcolor[HTML]{EFEFEF}Bldn & \cellcolor[HTML]{EFEFEF}Road & \cellcolor[HTML]{EFEFEF}Sdwlk & \cellcolor[HTML]{EFEFEF}Fence & \cellcolor[HTML]{EFEFEF}Vegittn & \cellcolor[HTML]{EFEFEF}Pole & - \\
\midrule
3D MinkNet14 \cite{choy20194d} & voxel & 1 & - & 89.39 & 97.68 & 69.43 & 86.52 & 98.11 & 97.26 & - \\
4D MinkNet14 \cite{choy20194d} & voxel & 3 & - & 90.13 & 98.26 & 73.47 & 87.19 & 99.10 & 97.50 & - \\
\midrule
PointNet++ \cite{qi2017pointnet++}  & point & 1 & -          & 96.88 & 97.72 & 86.20 & 92.75 & 97.12 & 97.09 & - \\
MeteorNet-m \cite{liu2019meteornet} & point & 2 & \checkmark & \textbf{98.22} & 97.79 & 90.98 & 93.18 & 98.31 & 97.45 & - \\
MeteorNet-m \cite{liu2019meteornet} & point & 2 & \xmark     & 97.65 & 97.83 & 90.03 & 94.06 & 97.41 & 97.79 & - \\
MeteorNet-l \cite{liu2019meteornet} & point & 3 & \xmark     & 98.10 & 97.72 & 88.65 & 94.00 & 97.98 & 97.65 & - \\
P4Transformer\cite{fan2021point}    & point & 1 & -      & 96.76 & 98.23 & 92.11 & 95.23 & 98.62 & 97.77 & - \\
P4Transformer\cite{fan2021point}    & point & 3 & \xmark & 96.73 & 98.35 & \textbf{94.03} & \textbf{95.23} & \textbf{98.28} & 98.01 & - \\
\midrule
MAMBA4D (ours) & point & 3 & \xmark & 96.16 & \textbf{98.58} & 92.80 & 94.95 & 97.08 & \textbf{98.24} & - \\
\midrule
\cellcolor[HTML]{EFEFEF}Method & \cellcolor[HTML]{EFEFEF}Input & \cellcolor[HTML]{EFEFEF}Frame & \cellcolor[HTML]{EFEFEF}Track & \cellcolor[HTML]{EFEFEF}Car & \cellcolor[HTML]{EFEFEF}T. Sign & \cellcolor[HTML]{EFEFEF}Pedstrn & \cellcolor[HTML]{EFEFEF}Bicycl & \cellcolor[HTML]{EFEFEF}Lane & \multicolumn{1}{c|}{\cellcolor[HTML]{EFEFEF}T. Light} & \cellcolor[HTML]{EFEFEF}mIoU \\
\midrule
3D MinkNet14 \cite{choy20194d}      & voxel & 1 & -          & 93.50 & 79.45 & 92.27 & 0.00 & 44.61 & \multicolumn{1}{c|}{66.69} & 76.24 \\
4D MinkNet14 \cite{choy20194d}      & voxel & 3 & -          & 94.01 & 79.04 & 92.62 & 0.00 & 50.01 & \multicolumn{1}{c|}{68.14} & 77.46 \\
\midrule
PointNet++ \cite{qi2017pointnet++}  & point & 1 & -          & 90.85 & 66.87 & 78.64 & 0.00 & 72.93 & \multicolumn{1}{c|}{75.17} & 79.35 \\
MeteorNet-m \cite{liu2019meteornet} & point & 2 & \checkmark & 94.30 & 76.35 & 81.05 & 0.00 & 74.09 & \multicolumn{1}{c|}{75.92} & 81.47 \\
MeteorNet-m \cite{liu2019meteornet} & point & 2 & \xmark     & 94.15 & 82.01 & 79.14 & 0.00 & 72.59 & \multicolumn{1}{c|}{77.92} & 81.72 \\
MeteorNet-l \cite{liu2019meteornet} & point & 3 & \xmark     & 93.83 & \textbf{84.07} & 80.90 & 0.00 & 71.14 & \multicolumn{1}{c|}{77.60} & 81.80 \\
P4Transformer\cite{fan2021point}    & point & 1 & -          & 95.46 & 80.75 & 85.48 & 0.00 & 74.28 & \multicolumn{1}{c|}{74.22} & 82.41 \\
P4Transformer\cite{fan2021point}    & point & 3 & \xmark     & 95.60 & 81.54 & \textbf{85.18} & 0.00 & 75.95 & \multicolumn{1}{c|}{79.07} & 83.16 \\
\midrule
MAMBA4D (ours) & point & 3 & \xmark & \textbf{95.75} & 82.03 & 84.57 & 0.00 & \textbf{79.35} & \multicolumn{1}{c|}{\textbf{80.74}} & \textbf{83.35} \\
\bottomrule

\end{tabular}
}
\vspace{-4mm}
\end{table*}

\subsection{Ablation Studies}
\label{sec:ablation}
In this section, we conduct ablation studies in terms of the effectiveness of Mamba modules, the number of Mamba blocks, positional encoding, and various scanning strategies. Notably, we conduct all the experiments here based on the baseline P4Transformer \cite{fan2021point} with 24-frame inputs.

\vspace{3pt}
\noindent\textbf{Spatial and temporal modeling.} Our baseline P4Transformer encodes the spatial and temporal embeddings with a Point 4D Convolution layer and a transformer layer, respectively. In Table \ref{tab:ablation1}, we separately remove our spatial and temporal Mamba modules to demonstrate the effectiveness of our proposed method. 

\begin{table*}[!t]
\resizebox{\linewidth}{!}{%
   
        \begin{minipage}[t]{0.24\textwidth}
            \centering
            \caption{Ablation studies of the intra-frame spatial Mamba and inter-frame temporal Mamba.}
            \label{tab:ablation1}
            \resizebox{\linewidth}{!}{
            \begin{tabular}{c|c|c}
                \toprule
                \cellcolor[HTML]{EFEFEF}Intra- & \cellcolor[HTML]{EFEFEF}Inter- &  \cellcolor[HTML]{EFEFEF}Accuracy (\%) \\
                \midrule
                \xmark &  \xmark & 90.94 \\
                \checkmark & \xmark & 91.36 \\
                \xmark & \checkmark & 91.67 \\
                \checkmark& \checkmark& \textbf{92.68} \\ 
                \bottomrule
            \end{tabular}
            }
        \end{minipage}
        \hspace{1em}
       
         \begin{minipage}[t]{0.24\textwidth}
            \centering
            \caption{Ablation studies of the number of Mamba blocks of intra- and inter-frame Mamba.}
            \label{tab:ablation2}
            \resizebox{\linewidth}{!}{
            \begin{tabular}{c|c|c}
                \toprule
                \cellcolor[HTML]{EFEFEF}Intra- & \cellcolor[HTML]{EFEFEF}Inter- &  \cellcolor[HTML]{EFEFEF}Accuracy (\%) \\
                \midrule
                4 & 4 & 90.24 \\
                4 & 12 & 89.89 \\
                12 & 4 & \bf{92.68} \\
                12& 12& 91.36 \\ \bottomrule
            \end{tabular}
            }
        \end{minipage}
        \hspace{1em}
       
        \begin{minipage}[t]{.18\linewidth}
            \centering
            \caption{Ablation studies of Position Encoding in Intra-frame Mamba.}
            \label{tab:ablation3}
            \resizebox{\linewidth}{!}{
            \begin{tabular}{c|c}
                \toprule
                \cellcolor[HTML]{EFEFEF}PE &  \cellcolor[HTML]{EFEFEF}Accuracy (\%) \\
                \midrule
                 \xmark & 89.20 \\
                3D  & 91.98 \\
               
                4D & \bf{92.68} \\
              \bottomrule
            \end{tabular}
            }
        \end{minipage}
        \hspace{1em}
      
        \begin{minipage}[t]{.18\linewidth}
            \centering
            \caption{Ablation studies of Position Encoding in Inter-frame Mamba.}
            \label{tab:ablation4}
            \resizebox{\linewidth}{!}{
            \begin{tabular}{c|c}
                \toprule
                \cellcolor[HTML]{EFEFEF}PE &  \cellcolor[HTML]{EFEFEF}Accuracy (\%) \\
                \midrule
                 \xmark & 88.85 \\
                3D  & 91.98 \\
        
                4D & \bf{92.68} \\
              \bottomrule
            \end{tabular}
            }
        \end{minipage}
      
}
\end{table*}

\begin{table*}[!t]
        \hspace{1em}
        \begin{minipage}[t]{.29\linewidth}
            \centering
            \caption{Ablation studies of different spatial ordering strategies in Intra-frame Mamba.}
            \label{tab:ablation5}
            \resizebox{\linewidth}{!}{
            \begin{tabular}{c|c}
             \toprule
            \cellcolor[HTML]{EFEFEF} Ordering & \cellcolor[HTML]{EFEFEF} Accuracy (\%) \\
             \midrule
X&88.50\\
Y& 88.50 \\
Z& 88.50 \\
XYZ & 87.80 \\
XZY & 87.45 \\
YXZ & 88.15 \\
YZX& 88.85 \\
ZXY & 87.80 \\
ZYX & 88.50 \\
Unidirection& \bf{92.68} \\

\bottomrule
\end{tabular}
            }
        \end{minipage}
        \hspace{2em}
        \begin{minipage}[t]{.61\linewidth}
            \centering
            \caption{Ablation studies of different spatial-temporal ordering strategies in Inter-frame Temporal Mamba. 'Sequential' means the temporally sequential scanning. 'Cross' means the cross-temporal scanning. }
            \label{tab:ablation6}
            \resizebox{\linewidth}{!}{
            \begin{tabular}{l|c|c|l|c|c}
\toprule
\multicolumn{2}{c|}{\cellcolor[HTML]{EFEFEF}{Ordering}} & \cellcolor[HTML]{EFEFEF} Accuracy (\%) & \multicolumn{2}{c|}{\cellcolor[HTML]{EFEFEF}{Ordering}} & \cellcolor[HTML]{EFEFEF} Accuracy (\%)\\
\midrule
&X& 89.55 &  &X& 89.55\\
&Y& 88.85 &&Y& 88.50 \\
&Z& 89.55 &&Z& 88.50 \\
&XYZ & 89.90 &&XYZ  & 89.90 \\
&XZY& 90.59 &&XZY& \textbf{92.68}\\
&YXZ &90.59 & &YXZ & \textbf{92.68}\\
&YZX& 90.09 &&YZX& 90.59\\
&ZXY & 90.24 &&ZXY & 92.33 \\
&ZYX &90.59& &ZYX & \textbf{92.68}\\
\multirow{-10}{*}{\begin{tabular}[c]{@{}c@{}} \rotatebox{90}{Sequential}\end{tabular}}&Unidirection& 91.36 &\multirow{-10}{*}{\begin{tabular}[c]{@{}c@{}} \rotatebox{90}{Cross}\end{tabular}}&-& - \\

\bottomrule
\end{tabular}
            }
        \end{minipage}
    \vspace{-5mm}
 
\end{table*} 

\begin{figure}[t]
\centering
\includegraphics[width=0.8\linewidth]{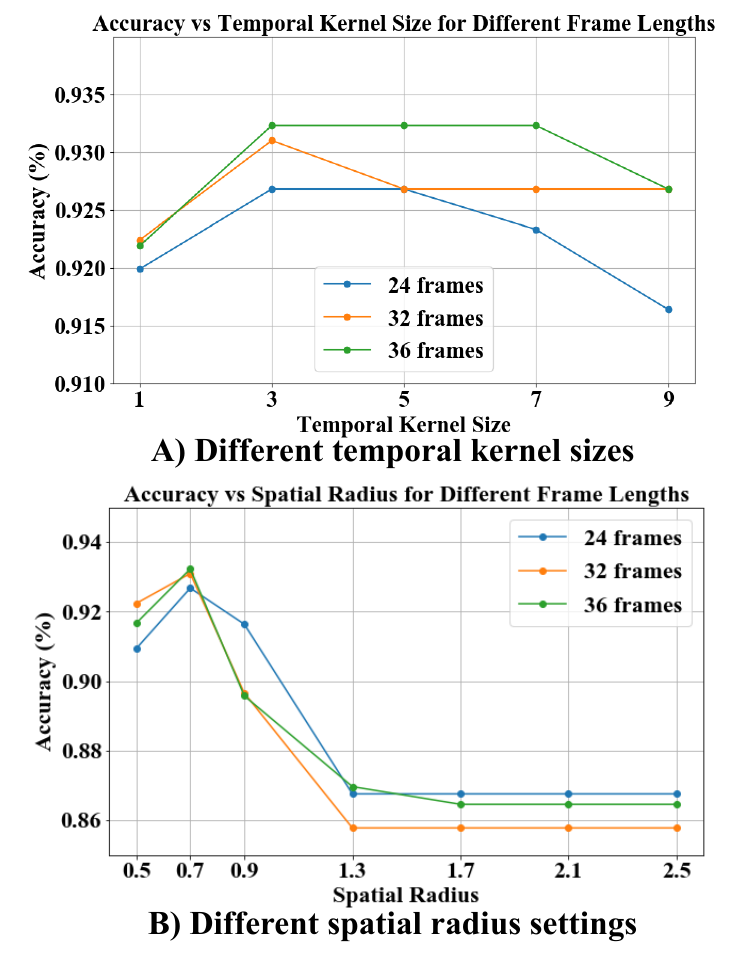}
\vspace{-4mm}
\caption{\textbf{Ablations of temporal stride and spatial radius settings.} We compare different temporal strides and spatial radius settings in creating spatio-temporal point tubes.}
\vspace{-6mm}
\label{fig:plot}
\end{figure}

\vspace{3pt}
\noindent\textbf{Number of blocks in Mamba modules.} Our spatio-temporal modeling is decoupled to an Intra-frame Spatial Mamba module and an Inter-frame Temporal Mamba module. Similar to transformer architectures, different Mamba blocks provide various network depths and modeling abilities. Here, we compare different Mamba block \cite{liang2024pointmamba} settings in Table \ref{tab:ablation2}.

\vspace{3pt}
\noindent\textbf{4D position encoding.} We encode 4D coordinates as additional inputs of Mamba blocks. We conduct studies with no position encoding or 3D spatial-only encoding in Table \ref{tab:ablation3} and Table \ref{tab:ablation4}. Experiment results demonstrate the effectiveness of our introduced 4D encoding method.

\vspace{3pt}
\noindent\textbf{Temporal stride and spatial radius settings.} We also conduct ablation studies about different temporal strides and the spatial radius settings in Fig. \ref{fig:plot}. From the table, the recognition accuracy achieves the best when the temporal kernel size is set as 3 and spatial radius is set as 0.7.

\vspace{3pt}
\noindent\textbf{Spatial scanning ordering in Intra-frame Mamba.}  We compare different orderings for the Intra-frame Mamba block in Table \ref{tab:ablation5}.

\vspace{3pt}
\noindent\textbf{Spatio-temporal scanning ordering in Inter-frame Mamba.} To effectively capture the spatio-temporal correlation for 4D videos, we design two main scanning strategies: Temporally sequential ordering and Cross-spatial ordering as in Section \ref{sec:inter}. In Table \ref{tab:ablation6}, we also compare unidirectional, one-time ordering (X, Y, Z), and three-time ordering (XYZ, XZY, YXZ, YZX, ZXY, ZYX) for each strategy. Among these, the cross-temporal ordering surpasses the other orderings, with a 92.68\% accuracy.

\section{Conclusion}
In this paper, we proposed a novel 4D backbone for the generic point cloud video understanding purely based on state space models. Decoupled spatial and temporal Mamba blocks are designed for the hierarchical spatio-temporal correlation. Short-term dynamics are established by the combination of an Intra-frame Spatial Mamba module and a temporal pooling layer. Another Inter-frame Temporal Mamba module is proposed to capture the long-term dynamics over the entire 4D video, where various scanning strategies are proposed. Extensive experiments on three 4D understanding tasks demonstrate the superiority of our Mamba-based 4D backbone and the perfect scalability to much larger input tokens with high efficiency. 


{
    \small
    \bibliographystyle{ieeenat_fullname}
    \bibliography{main}
}


\end{document}